\documentclass[11pt,a4paper]{article}
\PassOptionsToPackage{hyphens}{url}\usepackage{hyperref} 

\usepackage[hyperref]{computel3}
\usepackage{times}
\usepackage{latexsym}
\usepackage{graphicx} 
\usepackage{url}
\usepackage{todonotes}
\usepackage{soul}
\usepackage{mathdesign} 
\aclfinalcopy 
\usepackage{booktabs}

\usepackage{CJKutf8}

\usepackage{tikz}
\usepackage{pgfplots}

\definecolor{mygrey}{RGB}{229,229,229}
\definecolor{mygrey2}{RGB}{127,127,127}
\definecolor{mygrey3}{RGB}{240,240,240}

\pgfplotsset{
 	axis background/.style={fill=mygrey},
	tick style=mygrey2,
	tick label style=mygrey2,
	grid=both,
	xtick pos=left,
	ytick pos=left,
	tick style={
		major grid style={style=white,line width=1pt},minor grid style=mygrey3,
		tick align=outside,
	},
	minor tick num=1,
}

\makeatletter
\newcommand{\verbatimfont}[1]{\renewcommand{\verbatim@font}{\ttfamily#1}}
\makeatother

\title{User-Friendly Automatic Transcription of Low-Resource Languages: \\ Plugging ESPnet into Elpis}

\author{
  Oliver Adams\textsuperscript{ a }, Benjamin Galliot\textsuperscript{ b }, Guillaume Wisniewski\textsuperscript{ c }, Nicholas Lambourne\textsuperscript{ d e }, \\ \textbf{Ben Foley\textsuperscript{ d e }, Rahasya Sanders-Dwyer\textsuperscript{ d e }, Janet Wiles\textsuperscript{ d e }, Alexis Michaud\textsuperscript{ b },} \\ \textbf{Séverine Guillaume\textsuperscript{ b }, Laurent Besacier\textsuperscript{ f }, Christopher Cox\textsuperscript{ g }}, \\ \textbf{Katya Aplonova\textsuperscript{ h }, Guillaume Jacques\textsuperscript{ i }, Nathan Hill\textsuperscript{ j }} \\
  \textsuperscript{ a } Atos zData, United States of America \\
  \textsuperscript{ b } Langues et Civilisations à Tradition Orale (LACITO), CNRS-Sorbonne Nouvelle, France \\
  \textsuperscript{ c } Université Paris, Laboratoire de Linguistique Formelle (LLF), CNRS, France\\ 
  \textsuperscript{ d }The University of Queensland, Brisbane, Australia\\
  \textsuperscript{ e }ARC Centre of Excellence	for	the	Dynamics of Language (CoEDL), Australia\\
  \textsuperscript{ f }Laboratoire d'Informatique de Grenoble (LIG), CNRS-Université Grenoble Alpes, France \\
  \textsuperscript{ g }University of Alberta, Canada\\ 
  \textsuperscript{ h }Langage, Langues et Civilisation d'Afrique (LLACAN), CNRS-INALCO, France\\
  \textsuperscript{ i }Centre de Recherches Linguistiques sur l'Asie Orientale (CRLAO), CNRS-EHESS, France \\
  \textsuperscript{ j }School of Oriental and African Studies, University of London, United Kingdom \\
  {\tt oliver.adams@gmail.com, b.g01lyon@gmail.com,} \\
  {\tt guillaume.wisniewski@u-paris.fr,} \\
  {\tt \{n.lambourne|b.foley|uqrsand5|j.wiles\}@uq.edu.au,} \\
  {\tt \{alexis.michaud|severine.guillaume\}@cnrs.fr,}\\
  {\tt laurent.besacier@univ-grenoble-alpes.fr, cox.christopher@gmail.com,}\\
  {\tt \{aplooon|rgyalrongskad\}@gmail.com, nh36@soas.ac.uk}\\
 }

\date{}

\begin{document}

\maketitle

\begin{abstract}
   This paper reports on progress integrating the speech recognition toolkit ESPnet into Elpis, a web front-end originally designed to provide access to the Kaldi automatic speech recognition toolkit. The goal of this work is to make end-to-end speech recognition models available to language workers via a user-friendly graphical interface.
   Encouraging results are reported on (i)~development of an ESPnet recipe for use in Elpis, with preliminary results on data sets previously used for training acoustic models with the Persephone toolkit along with a new data set that had not previously been used in speech recognition, and (ii)~incorporating ESPnet into Elpis along with UI enhancements and a CUDA-supported Dockerfile.
\end{abstract}

\section{Introduction}


Transcription of speech is an important part of language documentation, and yet speech recognition technology has not been widely harnessed to aid linguists. Despite revolutionary progress in the performance of speech recognition systems in the past decade \cite{hinton2012deep,hannun2014deep,zeyer2018improved,hadian2018end,ravanelli2019pytorch,zhou2020rwth}, including its application to low-resource languages \cite{besacier_automatic_2014,blokland_language_2015,van_esch_future_2019,hjortnaes_towards_2020}, these advances are yet to play a common role in language documentation workflows. Speech recognition software often requires effective command line skills and a reasonably detailed understanding of the underlying modeling. People involved in language documentation, language description, and language revitalization projects (this includes, but is not limited to, linguists who carry out fieldwork) seldom have such knowledge. Thus, the tools are largely inaccessible by many people who would benefit from their use.

Elpis\footnote{\url{https://github.com/CoEDL/elpis}} is a tool created to allow language workers with minimal computational experience to build their own speech recognition models and automatically transcribe audio \cite{foley_building_2018,foley_elpis_2019}. Elpis uses the Kaldi\footnote{\url{https://github.com/kaldi-asr/kaldi}} automatic speech recognition (ASR) toolkit \cite{Povey_ASRU2011} as its backend. Kaldi is a mature, widely used and well-supported speech recognition toolkit which supports a range of hidden Markov model based speech recognition models.

In this paper we report on the ongoing integration of ESPnet\footnote{\url{https://github.com/espnet/espnet}} into Elpis as an alternative to the current Kaldi system. We opted to integrate ESPnet \cite{watanabe_espnet_2018} as it is a widely used and actively developed tool with state-of-the-art end-to-end neural network models. By supporting ESPnet in Elpis, we aim to bring a wider range of advances in speech recognition to a broad group of users, and provide alternative model options that may better suit some data circumstances, such as an absence of a pronunciation lexicon.

In the rest of this paper, we describe changes to the Elpis toolkit to support the new backend, and preliminary experiments applying our ESPnet recipe to several datasets from a language documentation context. Finally, we discuss plans going forward with this project.


\section{Related Work} 
\paragraph{Automatic phonetic/phonemic transcription in language documentation} As a subset of speech recognition research, work has been done in applying speech recognition systems to the very low-resource phonemic data scenarios typical in the language documentation context.  Encouraging results capitalizing on the advances in speech recognition technology for automatic phonemic transcription in a language documentation context were reported by \newcite{adams_evaluating_2018}. Their work used a neural network architecture with connectionist temporal classification \cite{Graves2006} for phonemic (including tonal) transcription. A command line toolkit was released called Persephone. To assess the reproducibility of the results on other languages, experiments were extended beyond the Chatino, Na and Tsuut'ina data sets, to a sample of languages from the Pangloss Collection, an online archive of under-resourced languages \cite{michailovsky_documenting_2014}. The results confirmed that end-to-end models for automatic phonemic transcription deliver promising performance, and also suggested that preprocessing tasks can to a large extent be automated, thereby increasing the attractiveness of the tool for language documentation workflows \cite{wisniewski_phonemic_2020}. Another effort in this space is Allosaurus \cite{allosaurus}, which leverages multilingual models for phonetic transcription and jointly models language independent phones and language-dependent phonemes. This stands as a promising step towards effective universal phonetic recognition, which would be of great value in the language documentation process.


\paragraph{User-friendly speech recognition interfaces} Since such research tools do not have user friendly interfaces, efforts have been put into making these tools accessible to wider audience of users. The authors of Allosaurus provide a web interface online.\footnote{\url{https://www.dictate.app}} To integrate Persephone into the language documentation workflow, a plugin, Persephone-ELAN,\footnote{\url{https://github.com/coxchristopher/persephone-elan}} was developed for ELAN,\footnote{\url{https://archive.mpi.nl/tla/elan}} a piece of software that is widely used for annotation in language documentation \cite{cox_persephone-elan_2019}.

Meanwhile, Elpis is a toolkit that provides a user-friendly front-end to the Kaldi speech recognition system. The interface steps the user through the process of preparing language recordings using existing ELAN transcription files, training a model and applying the model to obtain a hypothesis orthographic transcription for untranscribed speech recordings.


\begin{figure*}[th!]
  \centering 
  \includegraphics[width=16cm]{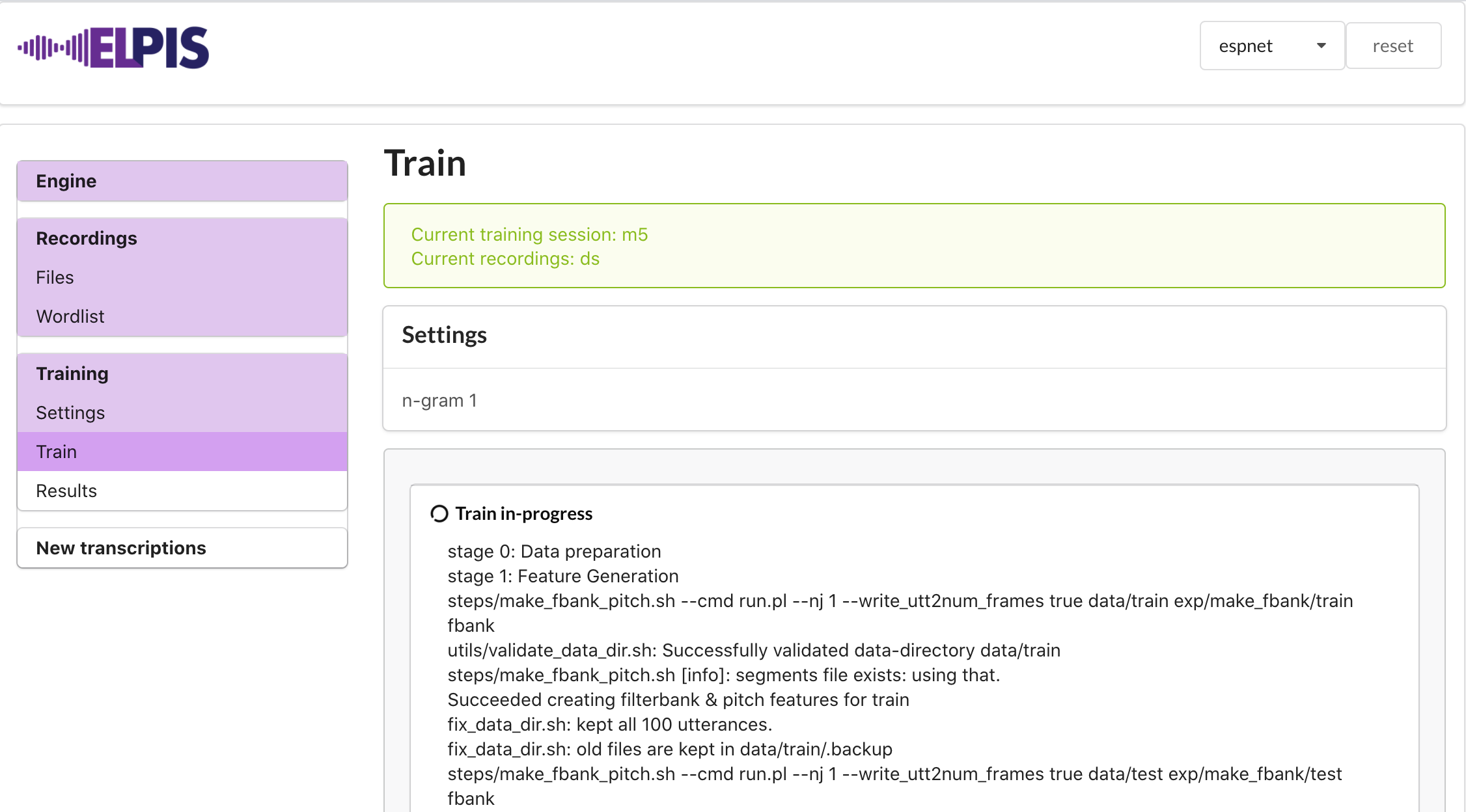}
  \caption{Training stages of the Elpis interface. Notice the choice of backend in the upper right-hand corner.}
  \label{fig:ElpisInterface}
\end{figure*}

\section{Bringing ESPnet to Elpis}

ESPnet is an end-to-end neural network-based speech recognition toolkit. Developed with Pytorch \cite{paszke2019pytorch} in a research context, the tool satisfies three desiderata for our purposes: (a) it is easy to modify training \emph{recipes}, which consist of collections of scripts and configuration files that make it easy to perform training and decoding by calling a wrapper script. These recipes describe a wide range of the hyperparameters and architecture choices of the model;
(b) it is actively developed, with frequent integration of the latest advances in end-to-end speech recognition; and (c) it supports Kaldi-style data formatting, which makes it a natural end-to-end counterpart to Kaldi backend that was already supported in Elpis. These points make it a more appealing candidate backend than Persephone, primarily due to ESPnet's larger developer base.

\subsection{Development of an ESPnet recipe for Elpis}
\label{sec:espnet_recipe}

One goal of the integration is to create a default ESPnet recipe for Elpis to use, that performs well across a variety of languages and with the small amount and type of data typically available in a language documentation context.

To get a sense of how easy it would be using ESPnet to get similar performance as previously attained we applied it to the single-speaker Na and Chatino datasets as used in \citet{adams_evaluating_2018} (see Table \ref{tab:results}, which includes other details of the datasets used, including the amount of training data). We report character error rate (CER) rather than phoneme error rate (PER) because it is general, does not require a subsequent language-specific post-processing step, and also captures characters that a linguist might want transcribed that aren't strictly phonemic. Because of minor differences in the training sets, their preprocessing, and metrics used, these numbers are not intended to be directly comparable with previous work. While these results are not directly comparable to the results they reported, the performance was good enough to confirm that integrating ESPnet was preferable to Persephone. We do no language-specific preprocessing, though the Elpis interface allows the user to define a character set for which instances of those characters will be removed from the text. For the Na data and the Japhug data in \S\ref{sec:japhug}, the  Pangloss XML format is converted to ELAN XML using a XSLT-based tool, Pangloss-Elpis.\footnote{\url{gitlab.com/lacito/pangloss-elpis}}

\begin{table*}
\begin{center}
\begin{tabular}{llllr}  
\toprule
Language    & Num speakers & Type & Train (minutes) & CER (\%) \\
\midrule
Na          & 1 & Spontaneous narratives & 273   & 14.5   \\
Na          & 1 & Elicited words \& phrases         & 188   & 4.7    \\
Chatino     & 1 & Read speech            & 81    & 23.5   \\
Japhug      & 1 & Spontaneous narratives & 170   & 12.8   \\
\bottomrule
\end{tabular}
\end{center}
\caption{Information on the evaluation datasets used and the character error rate performance of the current recipe.}
\label{tab:results}
\end{table*}

While we did not aggressively tune hyperparameters and architecture details, they do have a substantial impact on performance and computational requirements. Owing to the small datasets and limited computational resources of many of the machines that Elpis may run on, we used a relatively small neural network. In the future we aim to grow a representative suite of evaluation languages from a language documentation setting for further tuning to determine what hyperparameters and architecture best suit different scenarios. Though we aim for a recipe that does well across a range of possible language documentation data circumstances, the best architecture and hyperparameters will vary depending on the characteristics of the input dataset. Rather than have the user fiddle with such hyperparameters directly, which would undermine the user-friendliness of the tool, there is potential to automatically adjust the hyperparameters of the model on the basis of the data supplied to the model. For example, the parameters could be automatically set depending on the number of speakers in the ELAN file and the total amount of speech.



The architecture we used for these experiments is a hybrid CTC-attention model \cite{watanabe2017hybrid} with a 3-layer BiLSTM encoder and a single layer decoder. We use a hidden size of 320 and use an equal weighting between the CTC and attention objectives. For optimization we use a batch length of 30 and the Adadelta gradient descent algorithm \cite{zeiler2012adadelta}. For more details, we include a link to the recipe.\footnote{\url{https://github.com/persephone-tools/espnet/commit/1c529eab738cc8e68617aebbae520f7c9c919081}}

\subsection{Elpis enhancements}
\label{sec:elpis_enhancements}

Beyond integration of ESPnet into Elpis, several other noteworthy enhancements have been made to Elpis.

\paragraph{Detailed training feedback} Prior to the work reported in this paper, the progress of training and transcribing stages was shown as a spinning icon with no other feedback. Due to the amount of time it takes to train even small speech recognition models, the lack of detailed feedback may cause a user to wonder what stage the training was at, or whether a fault had caused the system to fail. During training and transcription, the backend processes' logs are now output to the screen (see Figure \ref{fig:ElpisInterface}). Although the information in these logs may be more detailed than what the intended audience of the tool needs to understand, it does serve to give any user feedback on how training is going, and reassure them that it \emph{is} still running (or notify them if a process has failed). The logs can also provide useful contextual information when debugging an experiment in collaborations between language workers and software engineers.

\paragraph{CUDA-supported Docker image} The type of Kaldi model which Elpis trains (using a Gaussian mixture model as the acoustic model) was originally selected to be computationally efficient, and able to run on the type of computers commonly used by language researchers, which often don't have a GPU (graphics processing unit). With the addition of ESPnet, the benefit of using a GPU will be felt through vastly reduced training times for the neural network. To this end, Elpis has been adapted to include Compute Unified Device Architecture (CUDA) support, which is essential in order to leverage a GPU when training ESPnet on a machine that has one available.




\section{Application to a new data set: Japhug}
\label{sec:japhug}

The point of this work is to provide a tool that can be used by linguists in their limited-data scenarios. To this end we aim to experiment with diverse datasets that reflect the breadth of language documentation contexts. Going forward, this will be useful in getting a sense of what sort of model performance users can expect given the characteristics of dataset. In this section we report on further application of the model underpinning the Elpis--ESPnet integration to another data set.


Japhug is a Sino-Tibetan language with a~rich system of consonant clusters, as well as flamboyant morphology. In Japhug, syllables can have initial clusters containing at most three consonants, and at most one coda \cite{jacques_japhug_2019}. Japhug does not have lexical tones. The language's phonological profile is thus very different from Na \cite[about which see][]{michaud2017} and Chatino \cite{cruz2011phonology,cruzetal2014,cavar2016}.

The data set comprises a total of about 30 hours of transcribed recordings of narratives, time-aligned at the level of the sentence \cite{macaire2020}, which is a huge amount in a language documentation context. The recordings were made in the course of field trips from the first years of the century until now, in a quiet environment, and almost all of a single speaker. Our tests on various data sets so far suggest that these settings (one speaker -- hence no speaker overlap -- and clean audio) are those in which performance is most likely to be good when one happens to be training an acoustic model from scratch.

The full data set is openly accessible online from the Pangloss Collection, under a Creative Commons license, allowing visitors to browse the texts, and computer scientists to try their hand at the data set.\footnote{Each text has a Digital Object Identifier, allowing for one-click access. Readers are invited to take a look: \url{https://doi.org/10.24397/pangloss-0003360}} The data collector's generous approach to data sharing sets an impressive example, putting into practice some principles which gather increasing support, but which are not yet systematically translated into institutional and editorial policies \cite{garellek_toward_2020}.

The dataset can be downloaded by sending a request to the Cocoon data repository, which hosts the Pangloss Collection. A script, \textit{retriever.py},\footnote{\url{https://gitlab.com/lacito/pangloss-elpis}} retrieves resources with a certain language name. Data sets can then be created in various ways, such as sorting by speaker (tests suggest that single-speaker models are a good way to start) and by genre, e.g.~excluding materials such as songs, which are a very different kettle of fish from ordinary speech and complicate model training. 

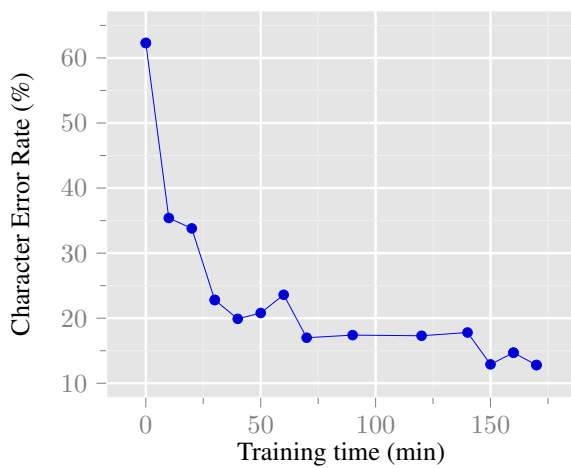
\begin{figure}[!ht]
  \begin{center}

    \begin{tikzpicture}[scale=.9]
      \begin{axis}[draw=white,
                   xlabel={Training time (min)},
                   ylabel={Character Error Rate (\%)}]
        \addplot coordinates {
          (0, 62.3) (0, 62.3) (10, 35.4)
          (20, 33.8) (30, 22.8) (30, 22.8)
          (40, 19.9) (50, 20.8) (60, 23.6)
          (70, 17.0) (90, 17.4) (120, 17.3)
          (140, 17.8) (150, 12.9) (160, 14.7)
          (160, 14.7) (170, 12.8) (170, 12.8)
        };
      \end{axis}
    \end{tikzpicture}

  \end{center}
  \vspace{-0.5cm}
  \caption{Character error rate for Japhug as a function of the amount of training data, using the ESPnet recipe included in Elpis.}
  \label{fig:ESPnet_Japhug_TrainingTime}
\end{figure}

\begin{figure}[!ht]
  \begin{center}

    \begin{tikzpicture}[scale=.9]
      \begin{axis}[draw=white,
                   legend entries={train,validation},
                   xlabel={Number of epochs},
                   ylabel={Character Error Rate (\%)}]
        \addplot coordinates {
          (1,  .991) (2, .836) (3, .643)
          (4,  .402) (5, .311) (6, .288)
          (7,  .231) (8, .205) (9, .190)
          (10, .181) (11, .176) (12, .161)
          (13, .158) (14, .142) (15, .133)
          (16, .122) (17, .118) (18, .103)
          (19, .091) (20, .080)
        };

        \addplot coordinates {
          (1,  .894) (2, .740) (3, .641)
          (4,  .301) (5, .329) (6, .286)
          (7,  .192) (8, .198) (9, .232)
          (10, .193) (11, .191) (12, .165)
          (13, .191) (14, .165) (15, .133)
          (16, .095) (17, .165) (18, .165)
          (19, .165) (20, .165)
        };

      \end{axis}
    \end{tikzpicture}

\end{center}
\vspace{-0.5cm}
\caption{Character error rate on the training set (blue) and validation set (orange) for Japhug as training progresses (up to 20 epochs), using the ESPnet recipe included in Elpis.}
\label{fig:ESPnet_Japhug_error_rate}
\end{figure}
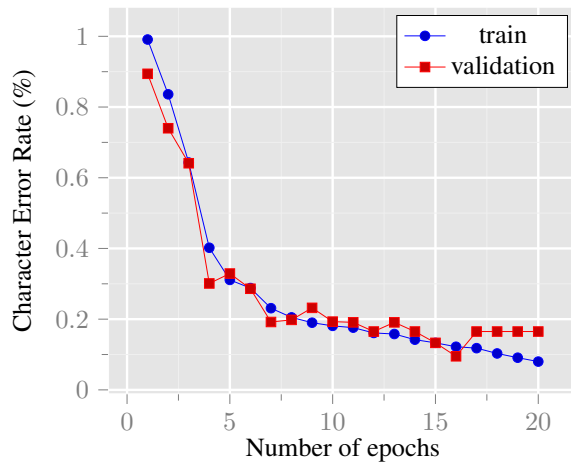

Figure~\ref{fig:ESPnet_Japhug_TrainingTime} shows how the phoneme error rate decreases as the amount of training data increases up to 170 minutes. Tests are currently being conducted to verify whether performance stagnates when the amount of data is increased beyond 170 minutes. As with the other experiments the recipe described in  \S\ref{sec:espnet_recipe} was used. For each amount of training data, the model was trained for 20 epochs for each of these training runs, with the smaller sets always as a subset of all larger sets. Figure \ref{fig:ESPnet_Japhug_error_rate} shows the training profile for a given training run using 170 minutes of data.

\section{Challenges concerning adoption of automatic speech recognition tools in language documentation}

Devoting a section to reflections about adoption of automatic speech recognition tools in language documentation may seem superfluous here. The audience of a conference on the use of computational methods in the study of endangered languages is highly knowledgeable about the difficulties and the rewards of interdisciplinary projects, as a matter of course. But it seemed useful to include a few general thoughts on this topic nonetheless, for the attention of the broader readership which we hope will probe into the Proceedings of the ComputEL-4 conference: colleagues who may consider joining international efforts for wider adoption of natural language processing tools in language documentation workflows. We briefly address a few types of doubts and misgivings.

\subsection{Is automatic speech recognition software too complex for language workers?}
A first concern is that automatic speech recognition software is simply too complex for language workers. But it should be recalled that new technologies that seem inaccessible to language workers can be game-changers in linguistics. For instance, the \LaTeX{} software is the typesetting backend used by the journal \textit{Glossa} \cite{rooryck_introducing_2016} and by the publishing house Language Science Press \cite{nordhoff_language_2018}, which publish research in linguistics, offering high-quality open-access venues with no author fees or reader fees. Thus, \LaTeX{}, a~piece of software which is notorious for its complexity, is used on a large scale in linguistics publishing: \textit{Glossa} publishes more than 100 articles a year, and Language Science Press about 30 books a year. Key to this success is an organizational setup whereby linguists receive not only a~set of stylesheets and instructions, but also hands-on support from a \LaTeX{} expert all along the typesetting process. Undeniably complex software is only accessible to people with no prior knowledge of it if support is available. Automatic speech recognition software should be equally accessible for language workers, given the right organization and setup. Accordingly, special emphasis is placed on user design in the Elpis project. This aspect of the work falls outside of the scope of the present paper, but we wanted to reassure potential users that it is clear to Elpis developers that the goal is to make the technology available to people who do not use the command line. If users can operate software such as ELAN then they will be more than equipped for the skills of uploading ELAN files to Elpis and clicking the Train button.

\subsection{Will the technology deliver on its promise?}

A second concern among language workers is whether the technology can deliver on its promise, or whether transcription acceleration projects are a case of “digital innovation fetishism” \cite{ampuja_blind_2020}. Some language workers have reported a feeling that integration of automatic transcription into the language documentation workflow \cite[as described in][]{michaud_integrating_2018} feels out of reach for them. There is no denying that natural language processing tools such as ESPnet and Kaldi are very complex, and that currently, the help of specialists is still needed to make use of this technology in language documentation. However, progress is clearly being made, and a motivated interdisciplinary community is growing at the intersection of language documentation and computer science, comprising linguists who are interested in investing time to learn about natural language processing and computer scientists who want to achieve “great things with small languages” \cite{thieberger_doing_2006}. It seems well worth investing in computational methods to assist in the urgent task of documenting the world's languages. 

\subsection{Keeping up with the state of the art vs.\ stabilizing the tool}

Finally, a~concern among linguists is that the state of the art in computer science is evolving so rapidly that the tool cannot be stabilized, and hence cannot be proposed to language workers for enduring integration into the language documentation workflow. In cases where significant, high-frequency updates are required to keep up with changes in speech recognition software, the investment could be too much for the relatively small communities of programmers involved in transcription acceleration projects. 

Our optimistic answer is that state-of-the-art code, or code close to the state of the art, need not be difficult to integrate, use or maintain. For example, the developers of Huggingface's Transformers \footnote{\url{https://github.com/huggingface/transformers}} do an impressive job of wrapping the latest and greatest in natural language processing into an easy-to-use interface \cite{Wolf2019HuggingFacesTS}. They have shown an ability to integrate new models quickly after their initial publication. Usability and stability of the interface is dictated by the quality of the code that is written by the authors of the backend library. If this is done well then the state of the art can be integrated with minimal coding effort by users of the library. For this reason, we are not so concerned about the shifting sands of the underlying building blocks, but the choice of quality backend library does count here. While it is true that there will have to be some modest effort to keep up to date with ESPnet (as would be the case using any other tool), in using ESPnet we are optimistic that the models supported by Elpis can remain up to date with the state of the art without too much hassle.

\section{Further improvements}
\label{sec:further_improvements}

The broader context to the work reported here is a rapidly evolving field in which various initiatives aim to package natural language processing toolkits in intuitive interfaces so as to allow a wider audience to leverage the power of these toolkits. Directions for new developments in Elpis include (i)~refining the ESPnet recipe, (ii)~refining the user interface through user design processes, (iii)~preparing pre-trained models that can be adapted to a small amount of data in a target language, and (iv)~providing Elpis as a web service.

\subsection{Refining the ESPnet recipe}

Refinement of the ESPnet recipe that is used in the Elpis pipeline, such that it works as well as possible given the type of data found in language documentation contexts, is a top priority. This work focuses on achieving lower error rates across data sets, starting with refining hyperparameters for model training and extends to other project objectives including providing pre-trained models (see \S\ref{sec:pretraining}). This work is of a more experimental nature and can be done largely independently of the Elpis front-end.

\subsection{Refining the interface}

In parallel with the technical integration of ESPnet with Elpis, a user-design process has been investigating how users expect to use these new features. In a series of sessions, linguists and language workers discussed their diverse needs with a designer. The feedback from this process informed the building of a prototype interface based on the latest version of Elpis at the time. The test interface was then used in individual testing sessions to discover points of confusion and uncertainty in the interface. Results of the design process will guide an update to the interface and further work on writing supporting documentation and user guides. The details of this process are beyond the scope of this paper and will be reported separately in future.

\subsection{Pre-trained models and transfer learning}
\label{sec:pretraining}

Adapting a trained model to a new language has a long history in speech recognition, having been used both for Hidden Markov Model based systems \cite{Schultz2001,le2005first,stolcke2006cross,toth2008cross,plahl2011cross,thomas2012multilingual,imseng2014using,do2014cross, heigold2013multilingual,scharenborg2017building} and end-to-end neural systems  \cite{toshniwal2017multilingual,chiu2017state,mueller2017phonemic,dalmia2018sequence,watanabe2017language,inaguma2018transfer,Yi2018,adams2019massively}. In scenarios where data in the target domain or language is limited, leveraging models trained on a number of speakers in different languages often can result in a better performance. The model can learn to cope with acoustic and phonetic characteristics that are common between languages, such as building robustness to channel variability due to different recording conditions, as well as learning common features of phones and sequences of phones between languages.

In recent years pre-training of models on large amounts of \emph{unannotated} data has led to breakthrough results in text-based natural language processing, initially gaining widespread popularity with the context-independent embeddings of word2vec \cite{Mikolov2013b} and GloVe \cite{Pennington2014}, before the recent contextual word embedding revolution \cite{peters2018deep,devlin2019bert,liu2020survey} that has harnessed the transformer architecture \cite{vaswani2017attention}. It is now the case that the best approaches in natural language processing are typically characterized by pre-training of a model on a large amount of unannotated data using the cloze task (a.k.a masked language model training) before fine-tuning to a target task. Models pre-trained in this way best make use of available data since the amount of unannotated data far outweighs annotated data and such pre-training is advantageous to downstream learning, whether a small or large amount of data is available in the target task \cite{gururangan2020dont}. Despite the established nature of pre-training in natural language processing, it is less well established in speech recognition, though there has been recent work \cite{riviere2020unsupervised,baevski2020wav2vec}.

The language documentation scenario, where annotated data is very limited is a scenario that we argue stands most to gain from such pre-training (both supervised and self-supervised out-of-domain); followed by model adaptation to limited target language data. One of the features Elpis could provide is to include pre-trained models in its distribution or via an online service. Such models may be pre-trained in a self-supervised manner on lots of untranscribed speech, trained in a supervised manner on transcribed speech in other languages, or use a combination of both pre-training tasks. In cases where the pre-trained model was trained in a supervised manner, there is scope to deploy techniques to reconcile the differences in acoustic realization between phonemes of different languages via methods such as that of Allosaurus \cite{allosaurus} which uses a joint model of language-independent phones and language-dependent phonemes. Providing a variety of pre-trained models would be valuable, since the best seed model for adaptation may vary on the basis of the data in the target language \cite{adams2019massively}.

A recognized problem in language documentation is that, owing to the transcription bottleneck, a large amount of unannotated and untranscribed data ends up in \emph{data graveyards} \cite{himmelmann2006language}: archived recordings that go unused in linguistic research. It is frequently the case that the vast majority of speech collected by field linguists is untranscribed. Here too, self-supervised pre-training in the target language is likely a promising avenue to pursue, perhaps in tandem with supervised pre-training regiments. For this reason, we are optimistic that automatic transcription will have a role to play in almost all data scenarios found in the language documentation context -- even when training data is extremely limited -- and is not just reserved for certain single-speaker corpora with consistently high quality audio and clean alignments with text. In the past one could plausibly argue that the limited amount of transcribed speech as training data is an insurmountable hurdle in a language documentation context, but that will likely not remain the case.

One of the next steps planned for Elpis is to allow for acoustic models to be exported and loaded. Beyond the immediate benefit of saving the trouble of training models anew each time, having a library of acoustic models available in an online repository would facilitate further research on adaptation of acoustic models to (i)~more speakers, and (ii)~more language varieties. Building universal phone recognition systems is an active area of research \cite{allosaurus}; these developments could benefit from the availability of acoustic models on a range of languages. Hosting acoustic models in an online repository, and using them for transfer learning, appear as promising perspectives.

\subsection{Providing Elpis as a web service}

Training models requires a lot of computing power. Elpis now supports high-speed parallel processing in situations where the user's operating system has compatible GPUs (see Section \ref{sec:elpis_enhancements}). However, many users don't have this technology in the computers they have ready access to, so we also plan to investigate possibilities for hosting Elpis on a high-capacity server for end-user access. Providing language technologies via web services appears to be a successful method of making tools widely available, with examples including the WebMAUS forced-alignment tool.\footnote{\url{https://clarin.phonetik.uni-muenchen.de/BASWebServices/interface}} The suite of tools provided by the Bavarian Speech Archive \cite{kisler_multilingual_2017} have successfully processed more than ten million media files since their introduction in 2012. For users who want to avoid sending data to a server, there are other possibilities: Kaldi can be compiled to Web Assembly so it can do decoding in a browser \cite{hu_kaldi-web_2020}. But for the type of user scenarios considered here, hosting on a server would have major advantages, and transfer over secure connection is a strong protection against data theft (for those data sets that must not be made public, to follow the consultants' wishes or protect the data collectors' exclusive access rights to the data so that they will not be scooped in research and placed at a disadvantage in job applications). 

This context suggests that it would be highly desirable to design web hosting for Elpis. It would facilitate conducting broad sets of tests training acoustic models, and would also facilitate the transcription of untranscribed recordings. 

\section{Conclusion}

In this paper we have reported on integrating ESPnet, an end-to-end neural network speech recognition system, into Elpis, the user-friendly speech recognition interface. We described changes that have been made to the front-end, the addition of a CUDA supported Elpis Dockerfile, and the creation of an ESPnet recipe for Elpis. We reported preliminary results on several languages and articulated plans going forward.

\section*{Acknowledgments}

Many thanks to the three reviewers for comments and suggestions. 

We are grateful for financial support to the Elpis project from the Australian Research Council Centre of Excellence for the Dynamics of Language, the University of Queensland, the \textit{Institut des langues rares} (ILARA) at \textit{École Pratique des Hautes Études}, the European Research Council (as part of the project “Beyond boundaries: Religion, region, language and the state” [ERC-609823]) and \textit{Agence Nationale de la Recherche} (as part of two projects, “Computational Language Documentation by 2025” [ANR-19-CE38-0015-04] and “Empirical Foundations of Linguistics” [ANR-10-LABX-0083]). 

Linguistic resources used in the present study were collected as part of projects funded by the European Research Council (“Discourse reporting in African storytelling” [ERC-758232]) and by \textit{Agence Nationale de la Recherche} (“Parallel corpora in languages of the Greater Himalayan area” [ANR-12-CORP-0006]).



\bibliography{computel3}
\bibliographystyle{acl_natbib_nourl}

\end{document}